\documentclass[letterpaper]{article} 
\usepackage{aaai25}  
\usepackage{times}  
\usepackage{helvet}  
\usepackage{courier}  
\usepackage[hyphens]{url}  
\usepackage{graphicx} 
\usepackage{booktabs}
\usepackage{algorithm}
\usepackage{algorithmic}
\usepackage{subcaption}
\usepackage{natbib}  
\usepackage{caption} 
\usepackage{algorithm}
\usepackage{algorithmic}

\usepackage{newfloat}
\usepackage{listings}
\DeclareCaptionStyle{ruled}{labelfont=normalfont,labelsep=colon,strut=off} 
\floatstyle{ruled}
\newfloat{listing}{tb}{lst}{}
\floatname{listing}{Listing}
\title{CLLoRA: An Approach to Measure the Effects of the \\ 
Context Length for LLM Fine-Tuning}
\author{
    Ping Zhang\textsuperscript{\rm 1},
    Zhaorui Zhang\textsuperscript{\rm 1}\thanks{Zhaorui Zhang is the corresponding author.},
    Sheng Di\textsuperscript{\rm 2},
    Yao Xin\textsuperscript{\rm 3},
    Benben Liu\textsuperscript{\rm 4}
}
\affiliations{
    \textsuperscript{\rm 1}The Hong Kong Polytechnic University\\

    \textsuperscript{\rm 2}Argonne National Laboratory, USA\\
    \textsuperscript{\rm 3}Guangzhou University\\
    \textsuperscript{\rm 4}The University of Hong Kong\\
}

\usepackage{bibentry}

\begin{document}

\maketitle

\begin{abstract}
Large language model fine-tuning has been identified as an efficient approach to applying the pre-trained Large language models to other domains. To guarantee data privacy for different data owners, models are often fine-tuned in federated learning environments across different data owners, which often involve data heterogeneity issues and affect the fine-tuning performance. In addition, the length of the context for the training data has been identified as a major factor that affects the LLM's model performance.

To efficiently measure how the context length affects the LLM's model performance in heterogeneous federated learning environments, we propose CLLoRA. CLLoRA utilizes the parameter-efficient fine-tuning approach LoRA based on different kinds of LLMs with varying sizes as the fine-tuning approach to investigate whether the quality and length of contexts can serve as standards for measuring non-IID context. The findings indicate that an imbalance in context quality not only affects local training on clients but also impacts the global model's performance. However, context length has a minimal effect on local training but a more significant influence on the global model. These results provide insights into how context quality and length affect the model performance for LLM fine-tuning in federated learning environments.
\end{abstract}

%

\section{Introduction}
Large language models (LLMs) \cite{touvron2023llama, openai2023gpt} have achieved great success across a variety of linguistic tasks, including text generation, program synthesis, question answering, and mathematical problem-solving. Emerging pre-trained large language models are open-sourced nowadays. Such models are trained based on public data collected from the web and have achieved a great breakthrough for the general text generation and program synthesis tasks. Recent advancements in large language modeling significantly depended on unsupervised training utilizing extensive volumes of human-generated text, primarily derived from web sources or curated corpora \cite{zhao2023survey, togetherai}. The demand for public human-generated text data grew with the rapid increase of the scale of large language models due to the high accuracy requirements. To scale up the large language models and train them efficiently, they are typically trained according to the neural scaling laws \cite{kaplan2020scaling}. Such relationships indicate that increasing the size of the training datasets is significant for efficiently improving the performance of the large language models. However, the high-quality public data will exhausted within a few years in the future according to the estimation of the data stocks \cite{villalobosposition,longpre2024consent}.


To address the public data exhaustion issue, private data has attracted our attention recently. Fine-tuning the large language models in specified domains based on private data owned by different organizations, such as government and hospitals, has become a new direction to keep the large language model moving forward. This advancement is further enhanced by the remarkable zero-shot and few-shot learning capabilities of emerging foundation models. Existing LLMs, including the GPT \cite{achiam2023gpt, brown2020language} and PaLM series \cite{chowdhery2023palm, driess2023palm}, are trained on a vast and diverse array of predominantly unlabeled data, with parameter counts reaching up to hundreds of billions. This extensive training enables these models to be effectively applied across various domains, requiring only minimal additional training, such as fine-tuning, on the specific target datasets \cite{cho2023heterogeneous}.


Federated learning, also known as collaborative learning, is a sub-field of machine learning that focuses on training models in decentralized settings and guarantees the privacy of the training data. The term federated learning was introduced by Google in 2016 to solve the challenge of training a centralized machine learning model when data is distributed among millions of clients \cite{mcmahanCommunicationEfficientLearningDeep2023, zhangSurveyFederatedLearning2021}. In traditional machine learning, data is collected centrally, and a single model is trained on this combined dataset, which will bring privacy problems and communication costs. In federated learning, only the locally trained models or computed gradients are exchanged without exposing any data information. As a result, it is able to protect privacy to some extent \cite{liuRecentAdvancesFederated2023}, while it also raises some problems. Given the distributed nature, the statistics of the data across different devices are likely to differ significantly. In other words, data is non-independently and identically distributed (non-IID), which will influence the performance of the aggregated model \cite{zhuFederatedLearningNonIID2021}. Thus, fine-tuning the pre-trained large language models in federated learning environments becomes the best choice for applying the pre-trained emerging LLMs in specified domains based on private data.

While fully fine-tuning large language models — retraining every parameter — tends to yield optimal results, the sheer volume of parameters makes this process highly resource-intensive and impractical for direct application to downstream tasks \cite{houlsbyParameterEfficientTransferLearning2019}. Parameter Efficient Fine Tuning (PEFT) strategies, conversely, adjust a minimal subset of parameters, drastically cutting down on computational and storage demands yet delivering results on par with comprehensive fine-tuning. For example, Low-Rank Adaptation (LoRA) achieves or surpasses full fine-tuning benchmarks on datasets like WikiSQL \cite{zhongSeq2SQLGeneratingStructured2017}, MNLI-m \cite{wangGLUEMultiTaskBenchmark2018}, SAMSum \cite{gliwaSAMSumCorpusHumanannotated2019}, and fine-tuning a significantly smaller parameter set in GPT-3 175B \cite{huLoRALowRankAdaptation2021}. Incorporating PEFT into federated learning enables client-server communication with minimal parameter exchange, focusing only on a select few, thereby substantially reducing both communication overhead and computational requirements \cite{kimClientCustomizedAdaptationParameterEfficient2023}.

Due to the heterogeneous feature of training data across different clients for federated learning environments, it is crucial to identify appropriate metrics for evaluating the heterogeneity of text data. While numerous studies have focused on image-based and textual datasets, research on text generation remains relatively sparse. In this work, we will employ LoRA fine-tuning, specifically targeting text generation tasks, to investigate how the non-IID text influences the efficacy of large language model fine-tuning in federated learning environments. We identify two challenges to investigate how the Non-IID text influences the performance of the large language model fine-tuning in the federated learning environments.

\textit{Firstly,} previous works indicate only limited data is critical to achieving superior performance for the large language model training. However, identifying features for significant and critical data is challenging. We still lack a principled understanding of what kinds of data can make better contributions to the performance of the large language models.

\textit{Secondly,} for the image classification, Dirichlet distribution is often used to synthesize a population of Non-IID data across different clients in heterogeneous federated learning environments. However, research on the scenarios of text data often used for large language models remains relatively sparse. How to synthesize a Non-IID dataset for heterogeneous federated learning environments is challenging.

To address the above challenges, we propose \textit{CLLoRA}, an approach to measure the effects of the context length for LLM fine-tuning in federated learning systems. According to recent studies, Llama3 \cite{meta2024introducing} introduced a tokenizer with a vocabulary of 128K tokens, enabling more efficient language encoding than Llama2’s 32K vocabulary. Additionally, Llama3 increased its MLP intermediate size from 11k to 14k. These changes reflect a trend toward more extensive vocabulary and intermediate sizes for better quality. Meanwhile, Llama3 maintains its hidden size of 4k for inference efficiency. This trend is also reflected in the Microsoft development of Phi-3 \cite{abdin2024phi} compared with Phi-2 \cite{javaheripi2phi}. All this evidence indicates that the length of the context is one of the key factors that affect the model performance for the large language model fine-tuning. Thus, \textit{CLLoRA} divides the whole training dataset into several classes according to the length of the context and then utilizes the Dirichlet distribution to synthesize the dataset for different clients for large language models fine-tuning in federated learning environments. In addition, \textit{CLLoRA} also evaluates how the context length affects the training performance for the large language model fine-tuning in federated learning environments. 

We list our contributions as the following:

\begin{itemize}
    \item This is the first work to utilize the context length of the data to divide the whole training dataset into several classes to synthesize the dataset for different clients for large language models fine-tuning in federated learning environments.
    \item We provide an in-depth analysis of how the heterogeneous data affect the model performance for both the global model on the server and the local model on each client for the large language models fine-tuning in the federated learning environments.    
    \item We perform a comprehensive evaluation of \textit{CLLoRA} on the most popular large language models OPT-125M, OPT-350M, OPT-1.3B based on various datasets, including the ``textgenerator-ds-mini" in Non-IID scenarios. Evaluation results show that the heterogeneity of the context length and quantity affects the generalization ability and fairness across different clients extremely.
\end{itemize}

\section{Background and Related Works}



\subsection{Federated Learning Systems}

The federated learning system is an innovative machine learning paradigm that addresses the growing concerns of data privacy and security by enabling decentralized model training across multiple client devices without the need to share raw data \cite{baumgartNotAllFederated2024}. Unlike traditional centralized approaches, where data is aggregated into a central repository for model training, federated learning systems allow each client to train a local model using its own data and subsequently share only the model updates with a central server. This collaborative approach not only preserves data privacy but also leverages the computational power of edge devices, thereby reducing latency and enhancing scalability \cite{liFederatedOptimizationHeterogeneous}. Despite its numerous advantages, federated learning faces challenges such as handling heterogeneous data distributions, managing communication overhead, and ensuring the security and integrity of model updates \cite{karimireddySCAFFOLDStochasticControlled2020}. Lots of existing work focuses on data compression to reduce the communication overhead \cite{di2024survey,huang2024optimized,huang2023c}. As research in this field progresses, federated learning holds the potential to revolutionize various domains, including healthcare, finance, and smart devices, by enabling privacy-preserving and efficient machine learning solutions \cite{acarFederatedLearningBased2021a}.



\subsection{LLMs and PEFT Methods}
Language models that have undergone pre-training and possess a substantial number of parameters, along with comprehensive training datasets, are commonly referred to as Large Language Models (LLMs) \cite{raffelExploringLimitsTransfer2023, zhangOPTOpenPretrained2022, touvronLLaMAOpenEfficient2023}. Most LLMs are based on the transformer architecture, and their parameter count is generally from 6 to 10 billion. 


The parameter-efficient fine-tuning (PEFT) strategies, based on operational approaches, can be categorized into four main types: additive, selective, re-parameterised, and hybrid PEFT \cite{hanParameterEfficientFineTuningLarge2024}. 

\textit{Additive PEFT} enhances the model's architecture by adding new trainable components or parameters. Within this category, methods such as Serial Adapter and Parallel Adapter \cite{heUnifiedViewParameterEfficient2022} are prominent. The former integrates adapters behind the transformer module, while the latter aligns them parallel to the module. \textit{Selective PEFT} focuses on training a subset of the model's parameters without increasing them. An example is Diff Pruning, which employs a diff vector to adaptively prune the model during training, using a differentiable approximation to the L0-norm penalty to promote sparsity \cite{guoParameterEfficientTransferLearning2021}. \textit{Re-parameterised PEFT} involves creating a lower-dimensional representation of the original model parameters. LoRA, a method under this category, achieves this through low-rank factorization by splitting the attention layers' weight matrices into two smaller matrices, thus significantly reducing the number of parameters to be fine-tuned. Unlike LoRA, which uses a fixed rank, DyLoRA \cite{valipourDyLoRAParameterEfficient2023} dynamically adjusts the rank of the low-rank matrix during fine-tuning, depending on the task's complexity by modifying the matrix's dimensions. \textit{Hybrid Fine-tuning}, as the name suggests, combines different PEFT methods when commonalities are identified. UniPELT \cite{maoUniPELTUnifiedFramework2022}, for instance, supports various methods including Prefix-tuning \cite{liPrefixTuningOptimizingContinuous2021}, Adapter, LoRA, BitFit \cite{zakenBitFitSimpleParameterefficient2022}, and their combinations. It utilizes a gating mechanism to activate the appropriate sub-modules for the given data or task.



\subsection{Non-IID Data in Federated Learning}
Substantial research has been conducted on federated learning with non-IID data, where data heterogeneity significantly impacts the performance of the aggregated model. The study \cite{luFederatedLearningNonIID2024} identifies three factors affecting model performance: data distribution imbalance, heterogeneous data characteristics, and differences in data volume. \cite{hsuMeasuringEffectsNonIdentical2019} employs a Dirichlet distribution to create unbalanced labels across clients using the CIFAR-10 dataset to evaluate the FedAvg algorithm. \cite{liFederatedLearningNonIID2022a} conducts a comprehensive study of five federated learning algorithms and nine visual datasets to assess the impact of non-IID data, utilizing three partitioning strategies: label distribution skew, feature distribution skew, and quantity skew. Although not directly related to federated learning, \cite{shenRethinkingDataSelection2024} demonstrates the influence of length and diversity on accuracy during supervised fine-tuning using LLaMA-2-7B \cite{touvronLlamaOpenFoundation2023}. Similarly, \cite{liuWhatMakesGood2024} introduces DEITA, a framework that measures data across three dimensions, including complexity, quality, and diversity, and evaluates training outcomes using multiple metrics. Although most of them are not focused on federated learning systems, these works provide valuable insights into assessing textual data diversity.

\section{Problem Formulation}
\begin{figure}[ht]
\vspace{-8pt}
    \centering
    \includegraphics[width=0.9\columnwidth]{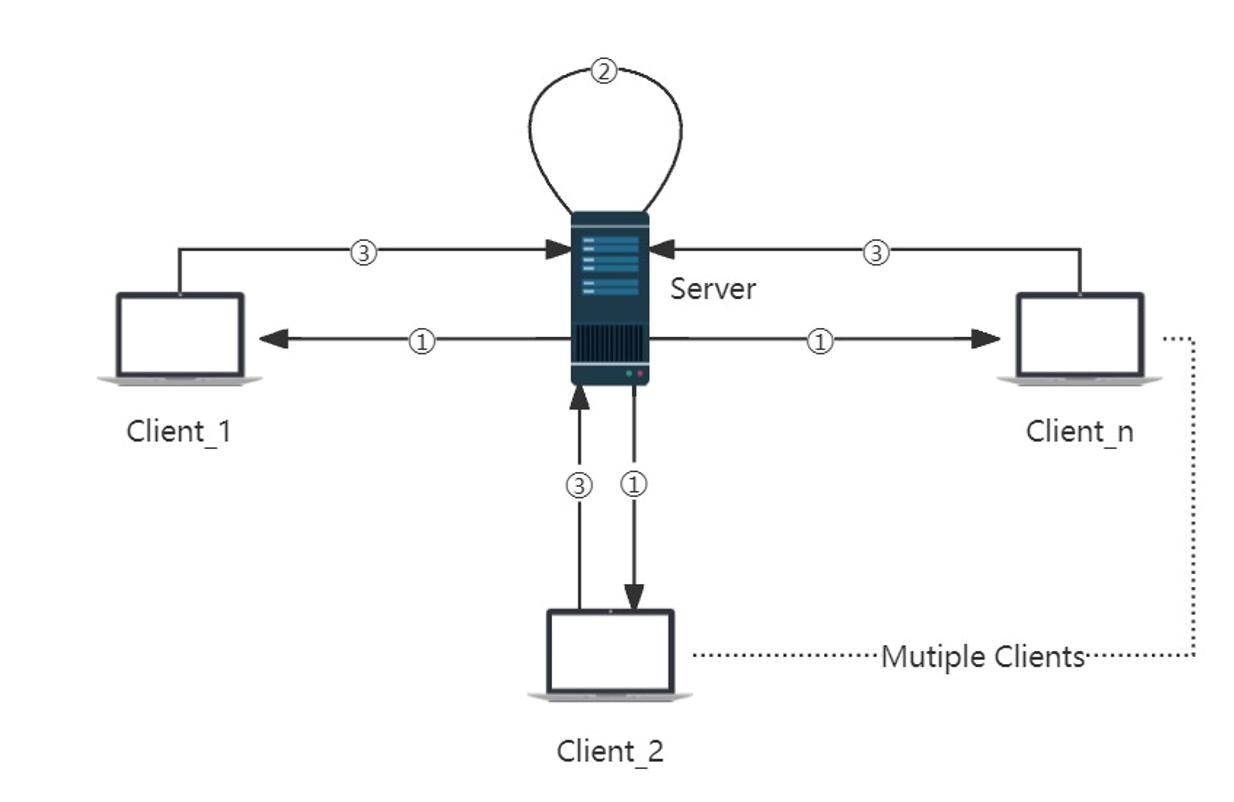}
    \caption{The process of horizontal federated learning.}
    \label{federated_learning}
    \vspace{-8pt}
\end{figure}

The process of horizontal federated learning is shown in Fig. \ref{federated_learning}. The process can be briefly described as follows: 1. the server dispatches the global model to a randomly chosen subset of clients; 2. these clients improve the model through training on their local datasets; 3. the clients then transmit their model parameters back to the server; 4. the server aggregates these parameters, enhancing the global model.


Data heterogeneity and communication overhead have been identified as two challenges for federated learning systems \cite{xu2024fedfa, zhang2022mipd, zhang2021sapus, zhang2022momentum}. Generally, full fine-tuning requires uploading all the model parameters, and then the communication overhead is the size of the whole model. With LoRA, the communication overhead will be positively correlated with the size of the set rank, so the communication overhead will be greatly reduced. There has been a lot of work exploring the impact of data heterogeneity between labeled images and labeled or prompt text in federated learning. Then, the question of how to define data heterogeneity in unlabelled and text-generation tasks is asked. Based on the introduction in previous sections, data heterogeneity has a greater impact on the learning effect. According to the FedAvg algorithm, we know that the uneven distribution of data quantity will affect the final integration effect of the model. The length of instruction in supervised fine-tuning tasks has a more obvious effect. Therefore, this work focuses on exploring the effect of text length on the aggregation effect of the model and exploring whether the length of the training text can be an influential factor. We will also use different sizes of models to see if the effect of data heterogeneity is generalized.

\section{The Design of \textit{CLLoRA}}

\subsection{Parameter-Efficient Fine-Tuning: LoRA}
LLMs often have an excess of parameters and only a minor fraction being crucial for particular tasks due to the vast parameter matrix. LoRA is designed based on the concept of employing matrix decomposition with lower dimensions for approximation purposes. By incorporating trainable matrices that decompose by rank into each transformer layer, LoRA effectively diminishes the number of parameters required for tasks that follow the initial training phase.

\begin{figure}[t]
    \centering
    \includegraphics[width=0.8\columnwidth]{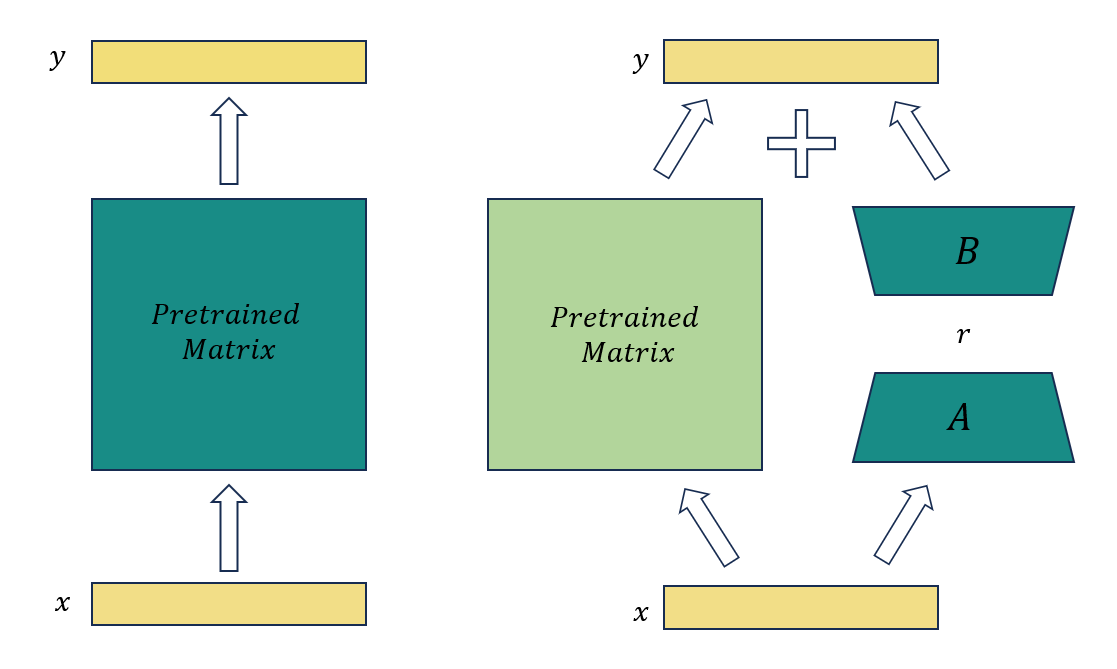}
    \caption{The design strategy of LoRA.}
    \vspace{-3pt}
    \label{LoRA}
    \vspace{-8pt}
\end{figure}

The specific approach involves adding a bypass structure to the network, where the bypass consists of the multiplication of two matrices, \(A\) and \(B\), as shown in Fig. \ref{LoRA}. The dimensions of the matrix \(A\) are \(m \times r\), and matrix \(B\) are \(r \times n\), where \(r\) is much smaller than \(min(m,n)\), typically taking values like 1, 2, 4, or 8. Consequently, the parameter count of this bypass is far less than the original network's parameters \(W\). During training, the original network's parameters \(W\) are frozen, and only the bypass parameters \(A\) and \(B\) are trained. Matrix \(A\) is initialized using a random normal distribution, while matrix \(B\) is set to zero. Since the parameter count for \(A\) and \(B\) is much less than \(W\), the memory overhead required for training is approximately equal to that during inference. 

\subsection{Implementing LoRA in Federated Learning} 
Typically, the FedAvg algorithm involves the server distributing the complete set of model parameters. After conducting local training, clients upload their entire model parameters back to the server for aggregation. In the case of using a large language model, LoRA is introduced for fine-tuning, and the uploaded parameters are those of the low-rank matrix. The implementation of FedAvg with LoRA introduces an additional hyperparameter: the matrix's rank, denoted as \(r\). Concurrently, the server's initialization model transitions to a pre-trained LLM along with $\Delta w_0$, as outlined in the pseudo-code of the Algorithm \ref{alg:fedavg_lora}.
\begin{algorithm}[t]
    \caption{Federated Learning System with LoRA}
    \label{alg:fedavg_lora}
    \begin{algorithmic}
    \REQUIRE $K$, the number of clients; $E$, the number of epochs; $\eta$, the learning rate; $C$, the fraction of clients selected per round; $r$, the rank of the low-rank decomposition
    \STATE Initialize global model weights $w_0$ and $\Delta w_0$
    \FOR{each round $t = 1,2,...,T$}
        \STATE $m \leftarrow \max(C \cdot K, 1)$
        \STATE $S_t \leftarrow$ (random set of $m$ clients)
        \FOR{each client $k \in S_t$ \textbf{in parallel}}
            \STATE $\Delta w_{t+1}^k \leftarrow$ ClientUpdate($k, \Delta w_t$)
        \ENDFOR
        \STATE $\Delta w_{t+1} \leftarrow \sum_{k=1}^{m} \frac{n_k}{n} \Delta w_{t+1}^k$ 
        \STATE $w_{t+1} \leftarrow w_0 + \Delta w_{t+1}$
    \ENDFOR
    
    \textbf{function} ClientUpdate($k, w$):
    \STATE Initialize local model with global weights $w$
    \FOR{each epoch $e = 1$ to $E$}
        \FOR{each batch $b$ in local dataset of client $k$}
            \STATE Compute gradient $\nabla \ell(\Delta w; b)$ with respect to the batch
            \STATE Update local model weights using gradient descent: $\Delta w \leftarrow \Delta w - \eta \nabla \ell(\Delta w; b)$
        \ENDFOR
    \ENDFOR
    \STATE \textbf{return} updated local model weights $\Delta w$
    \end{algorithmic}
\end{algorithm}

\subsection{Training Data Partitioning Strategies: Dirichlet}

\begin{figure*}
    \centering
    \includegraphics[width=0.96\linewidth, height=0.2\linewidth]{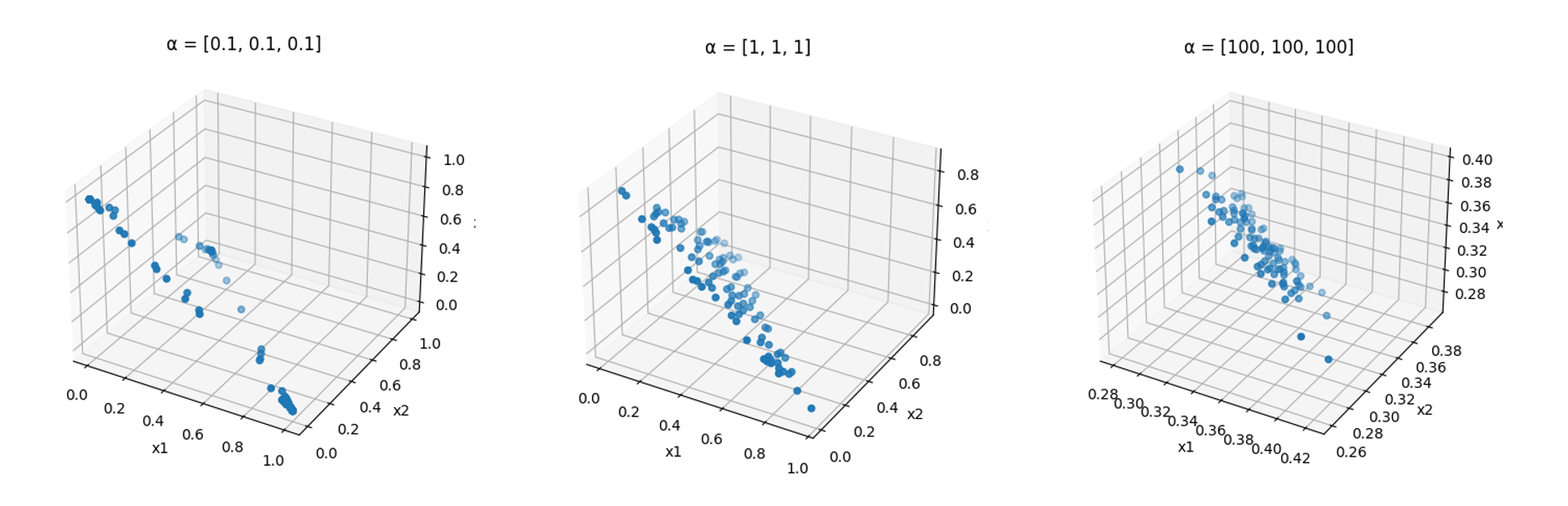}
    \caption{3D Scatter Plot of Dirichlet Distribution.}
    \label{fig:dirichlet}
\end{figure*}

The Dirichlet distribution constitutes a group of continuous multivariate probability distributions, which are parameterized by a vector of positive real numbers denoted as $\alpha$. This distribution finds extensive application within the domains of machine learning and natural language processing. The degree of data imbalance can be manipulated by varying the values of $\alpha$. A smaller $\alpha$ results in greater data imbalance. Fig. \ref{fig:dirichlet} shows the 3D scatter plot of different $\alpha$. 

\textit{CLLoRA} partition the training data set for different clients based on the Dirichlet distribution according to the following two factors: the length of the context and the number of sentences, also defined as the quantity of the sentences.

\subsubsection{Label Distribution Skew}
\begin{figure}[t]
    \centering
    \includegraphics[width=0.48\textwidth]{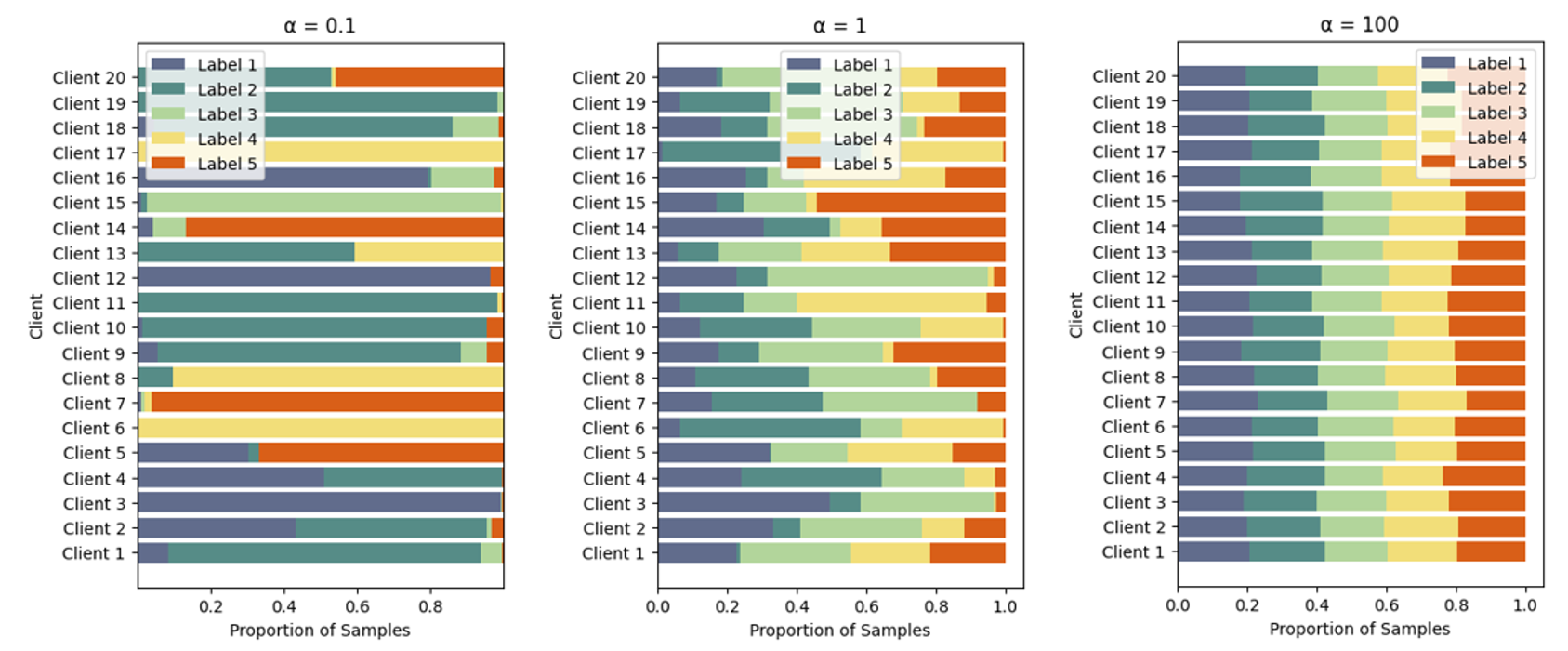}
    \caption{Different $\alpha$ for label distribution skew.}
    \label{clientlabels}
    \vspace{-8pt}
\end{figure}

The text data itself has no labels, and in order to explore whether different lengths of text affect the training effect, 5 labels were manually specified for different lengths of text. The effect of adjusting different $\alpha$, in the case of having 20 clients, is shown in Fig. \ref{clientlabels}. In this case, each client has an equal number of samples.

\subsubsection{Quantity Distribution Skew}

\begin{figure}[t]
    \centering
    \includegraphics[width=0.48\textwidth]{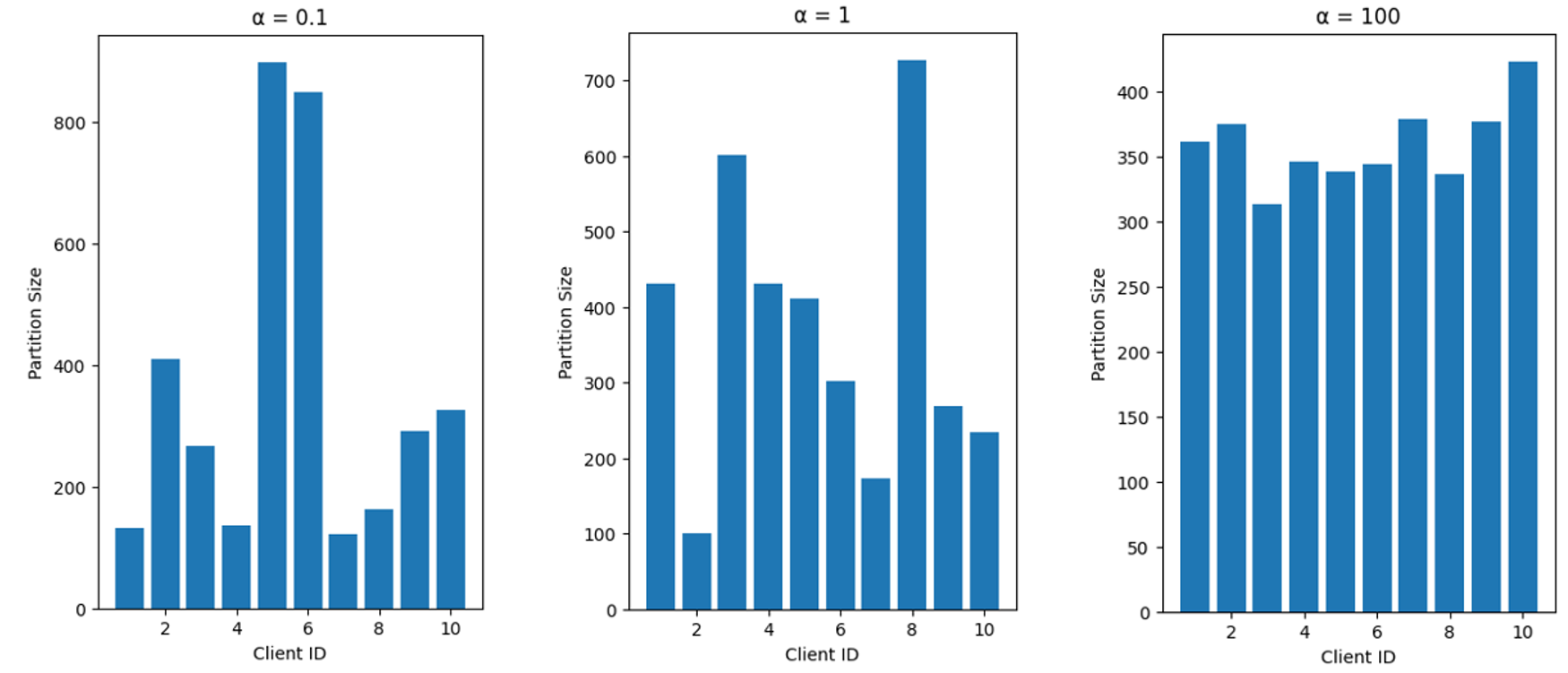}
    \caption{Different $\alpha$ for quantity distribution skew.}
    \label{clientquantity}
    \vspace{-8pt}
\end{figure}

In this case, the distribution of data length should be balanced, and the only difference is the amount of data. This is reflected in the text by the difference in the number of text entries in the dataset to which it is assigned. Using the Dirichlet distribution once again, the results are shown in Fig. \ref{clientquantity} for 10 clients. The server randomly selects clients for each communication round, and clients will train the model with local data. Then, the server gets the test loss of the aggregated model for each communication round. The experiment will compare the Global model loss curve with different $\alpha$.

\begin{table*}[t]
    \centering
    \fontsize{10}{12}\selectfont
    \begin{tabular}{ccccc}
    \toprule
    Model & Trainable Parameters & All Parameters & Ratio & Communication Cost\\ 
    \midrule
    OPT-125M & 73,728 & 125,313,024 & 0.0588\% & 0.30MB \\
    OPT-350M & 196,608 & 331,393,024 & 0.0593\% & 0.78MB \\
    OPT-1.3B & 393,216 & 1,316,151,296 & 0.0299\% & 1.53MB \\
    \bottomrule
    \end{tabular}   
    \caption{Comparison of different models after using LoRA.}
    \label{Parameters}
    \vspace{-8pt}
\end{table*}

\section{Experiment and Results Analysis}
\subsection{Prototype Implementation}
We implement our \textit{CLLoRA} on top of \textit{Plato} \cite{} and \textit{Pytorch}. Plato is a federated learning framework that supports temporal simulation for different kinds of training strategies running on a single machine. Our experiments were performed on an NVIDIA GeForce RTX 4090 graphic card.

\subsection{Evaluation Methodology}
\subsubsection{Benchmarks.} Experiments were conducted using three models of different popular LLMs: OPT-125M, OPT-350M, and OPT-1.3B, and dataset ``textgenerator-ds-mini". To simulate the Non-IID environments, we divide the whole training dataset into 5 classes according to the context length and the number of sentences. Then, we partition the dataset (with 5 classes) as $n$ parts ($n$ is the number of clients) based on the Dirichlet Distribution and use a coefficient $\alpha$ to control the heterogeneity of the data distribution across clients.


Because different-sized models have different learning abilities, different experiments were also set up with different \(E\). Comparisons after implementing LoRA with \(r = 2\) are shown in table \ref{Parameters}. It can be seen that the number of parameters and communication overheads have been reduced considerably. The reduction in the number of parameters also represents a lower requirement for memory and computational resources, optimizing the overall efficiency of the models. Despite the model being configured to truncate sequences with a maximum token length of 128, it is reasonable to infer that the truncation of text will influence the training outcomes. This is due to the varying degrees of information contained within the text. 

\subsubsection{Performance Metrics.}
We mainly focus on the analysis of the \textit{generalization ability} for the global model and the \textit{fairness} of the model performance across different clients, which is defined as the variance of the test loss for different clients. We analyze how the context length and the number of sentences affect the generalization ability for the global model and the fairness of the model performance across different clients in the following sections.

\subsection{Effect of Context Length on the Global Model}
\begin{figure}[t]
    \centering
    \begin{subfigure}[b] {0.46\textwidth}
        \includegraphics[width=\textwidth]{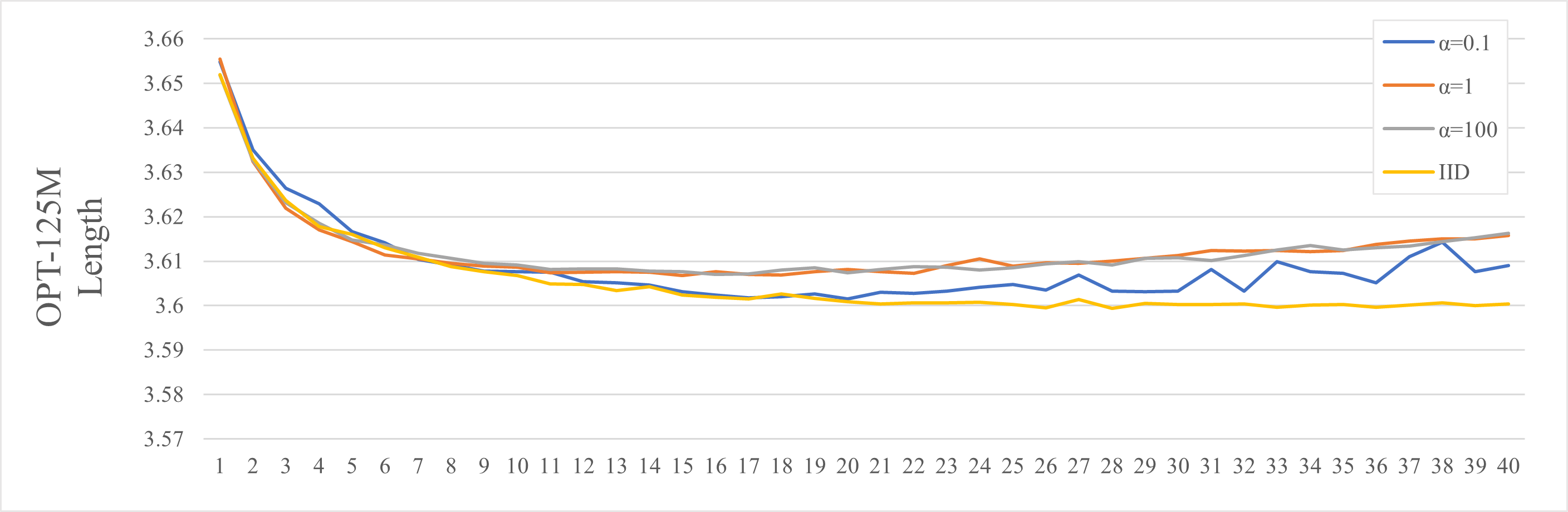}
        \caption{Test loss curves for OPT-125M, \(E\) = 5.}
        \label{fig:l1}
    \end{subfigure}
    
    \begin{subfigure}[b] {0.46\textwidth}
        \includegraphics[width=\textwidth]{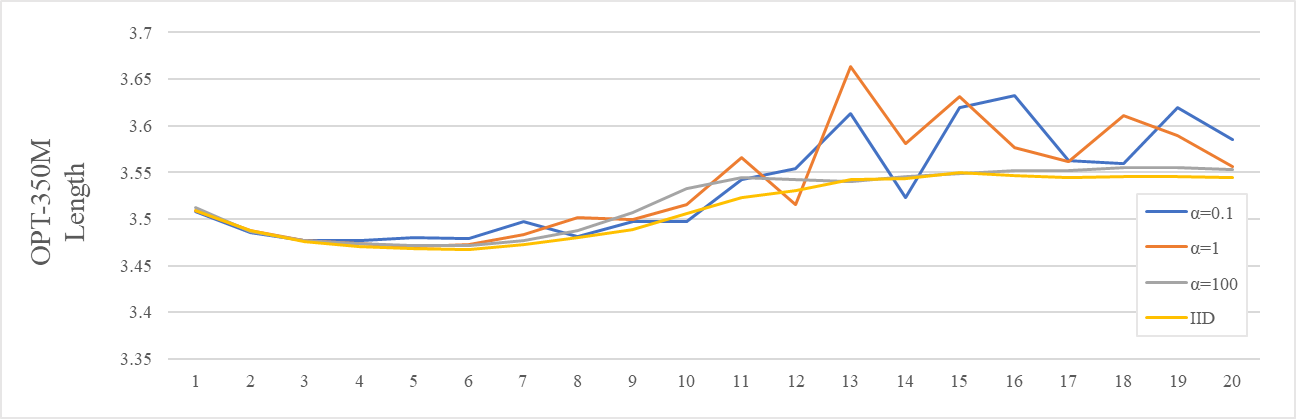}
        \caption{Test loss curves for OPT-350M, \(E\) = 3.}
        \label{fig:l2}
    \end{subfigure}
    
    \begin{subfigure}[b] {0.46\textwidth}
        \includegraphics[width=\textwidth]{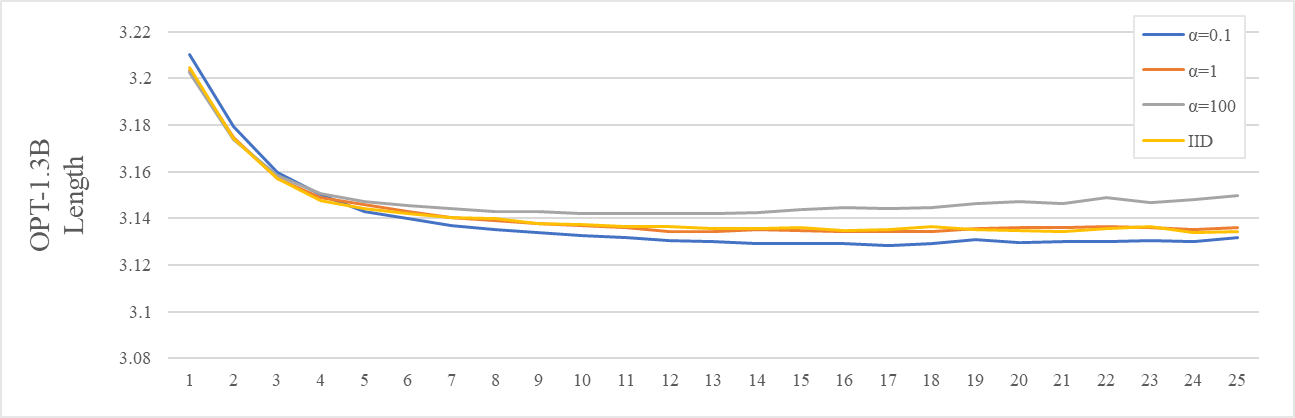}
        \caption{Test loss curves for OPT-1.3B, \(E\) = 1.}
        \label{fig:l3}
    \end{subfigure}
    
    \caption{The global model test loss of different models in the imbalanced control text length case. X-axis: communication round; Y-axis: test loss of the aggregated model.}
    \vspace{3pt}
    \label{fig:labels20c}
    \vspace{-8pt}
\end{figure}

In this experiment, the number of clients is 20. In every communication round, the server will select 2 clients and send the aggregated model parameters to selected clients. Label Distribution Skew describes the training results using different \(\alpha\) and IID data are shown in Fig. \ref{fig:labels20c}. 

In the OPT-125M model, each round of communication involves five epochs of local training. Notably, the test curve experiences significant fluctuations at \(\alpha\) = 0.1. As the rounds progress, a trend of overfitting becomes apparent in all test curves except for the IID. Transitioning to the OPT-350M model, the number of local training epochs per communication round decreases to $E=3$. At \(\alpha\) = 0.1 and 1, the curves exhibit noticeable fluctuations, and the extent of overfitting intensifies. Setting \(\alpha\) to 100 yields test curve variations that are relatively stable, akin to those of the IID, although overfitting occurs sooner. Regarding the OPT-1.3B model, despite each communication round comprising only one local training epoch, the curves are comparatively smoother due to the larger model's enhanced learning capacity. The curve patterns across all four tests display smoothness. In contrast to earlier models, the test loss reaches its minimum at \(\alpha\) = 0.1, whereas overfitting is detected at \(\alpha\) = 100. The curve at \(\alpha\) = 1 closely mirrors that of the IID. However, in the initial rounds, it is evident that at \(\alpha\) = 0.1, the performance is not as good as in other scenarios. This suggests that, in this context, the imbalance in text length actually enhances the model's generalization ability.

\subsection{Effect of Context Quantity on the Global Model}
\begin{figure}[t]
    \centering
    \begin{subfigure}[b]{0.46\textwidth}
        \includegraphics[width=\textwidth]{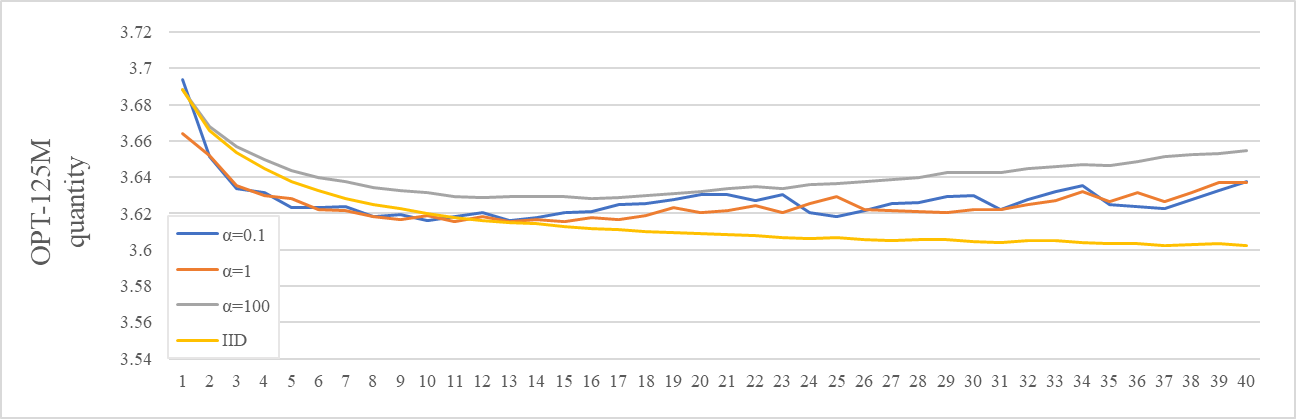}
        \caption{Test loss curves for OPT-125M, \(E\) = 5.}
        \label{fig:q1}
    \end{subfigure}
    
    \begin{subfigure}[b]{0.46\textwidth}
        \includegraphics[width=\textwidth]{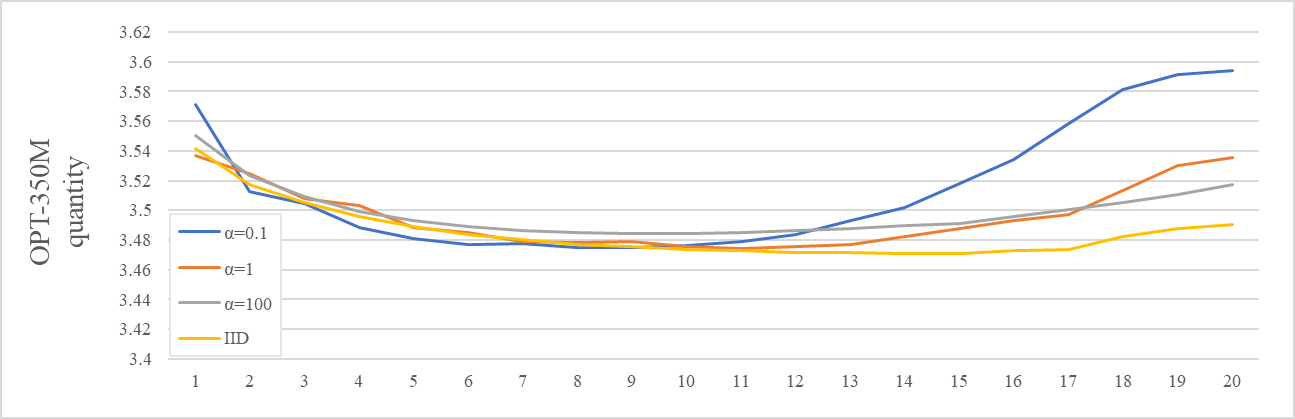}
        \caption{Test loss curves for OPT-350M, \(E\) = 3.}
        \label{fig:q2}
    \end{subfigure}
    
    \begin{subfigure}[b]{0.46\textwidth}
        \includegraphics[width=\textwidth]{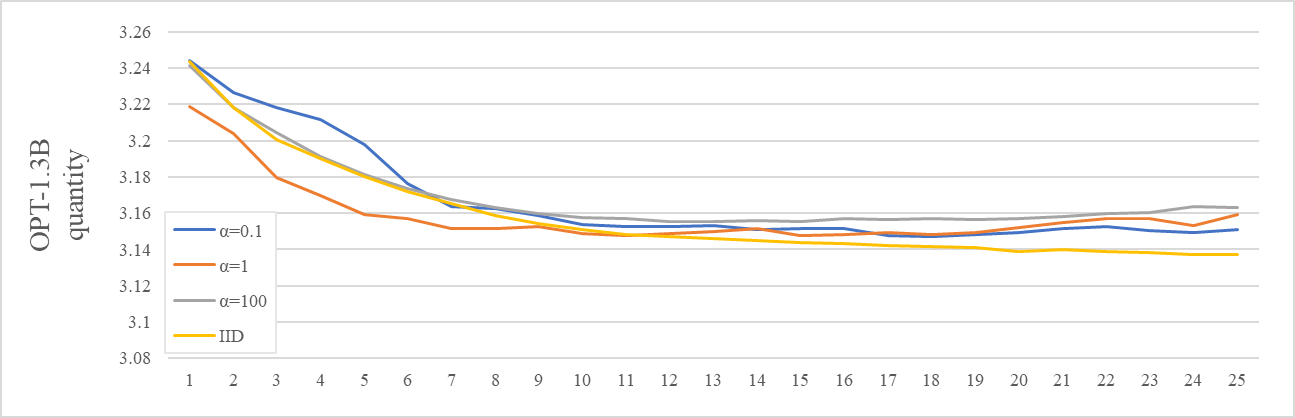}
        \caption{Test loss curves for OPT-1.3B, \(E\) = 1.}
        \label{fig:q3}
    \end{subfigure}
    
    \caption{The global model test loss of different models in the imbalanced control text quantity case. X-axis: communication round; Y-axis: test loss of the aggregated model.}
    \vspace{-3pt}
    \label{fig:quantity10c}
    \vspace{-8pt}
\end{figure}

In this experiment, the number of clients is 10. In every communication round, the server will select 2 clients and send the aggregated model parameters to selected clients. As described in Quantity Distribution Skew, the training results using different \(\alpha\) and IID data are shown in Fig. \ref{fig:quantity10c}.

For the OPT-125M model, the local training epoch is set to $E=5$ for every communication round. Fig. \ref{fig:quantity10c} reveals that the test curves for \(\alpha\) = 0.1 and 1 exhibit significant fluctuations, while the other two curves change more smoothly, indicating that the model is highly sensitive to imbalances in text quantity. The non-IID test curves show a clear trend of overfitting, with the curve for \(\alpha\) = 100 being the most pronounced. In contrast, the IID curve is relatively smooth and maintains a downward trend, with the test loss being the worst when \(\alpha\) = 100. Moving on to the OPT-350M model, the local training epoch for each communication round is $E=3$. All four test curves display overfitting tendencies, with \(\alpha\) = 0.1 showing the most noticeable trend. Although the overfitting trend for \(\alpha\) = 100 is not as apparent, it exhibits the worst performance in terms of test loss, whereas the IID scenario shows the best results. The OPT-1.3B model has only $E=1$ local training epoch per communication round. The test curves for \(\alpha\) = 0.1 and 1 have slight fluctuations, while the other two are more stable. Similar to the previous models, the IID scenario yields the best results, with the worst outcomes occurring at \(\alpha\) = 100. Across all three models, the test loss performance is poorest at \(\alpha\) = 100, while the most imbalanced distributions (corresponding to \(\alpha\) = 1 and 0.1) perform better, suggesting that extreme imbalance in distribution can enhance the model's generalization capabilities but may lower the performance ceiling.

\subsection{Effect of Context Length on Local Training}
\begin{figure}[t]
    \centering
    \begin{subfigure}[b]{0.2\textwidth}
        \includegraphics[width=\textwidth]{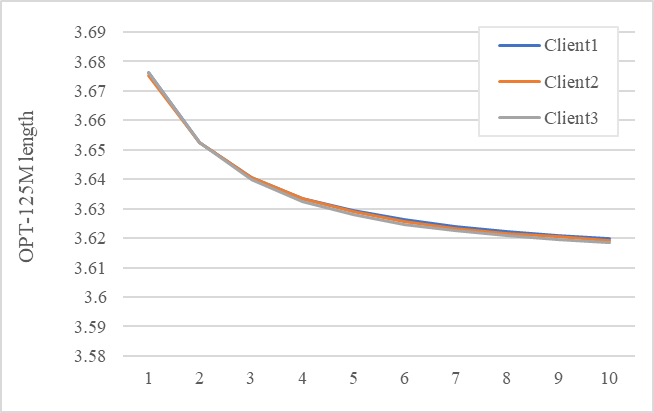}
        \caption{Test loss for OPT-125M with non-IID data. \(E\) = 3.}
        \label{fig:l11}
    \end{subfigure}
    \hspace{0.03\textwidth}
    \begin{subfigure}[b]{0.2\textwidth}
        \includegraphics[width=\textwidth]{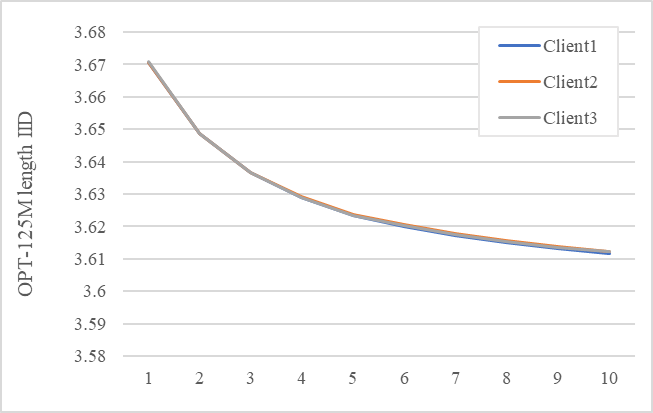}
        \caption{Test loss for OPT-125M with IID data. \(E\) = 3.}
        \label{fig:l12}
    \end{subfigure}

    \begin{subfigure}[b]{0.2\textwidth}
        \includegraphics[width=\textwidth]{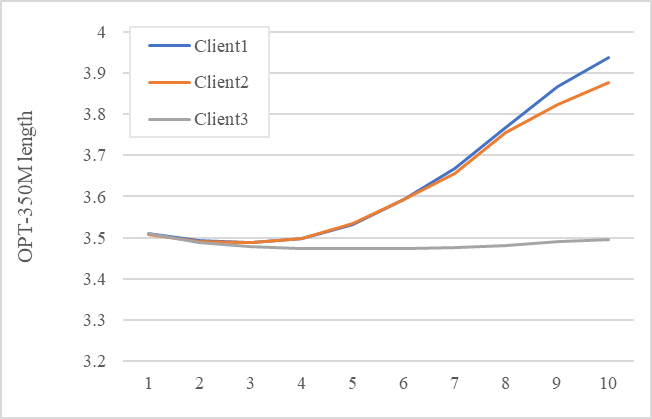}
        \caption{Test loss for OPT-350M with non-IID data. \(E\) = 3.}
        \label{fig:l21}
    \end{subfigure}
    \hspace{0.03\textwidth}
    \begin{subfigure}[b]{0.2\textwidth}
        \includegraphics[width=\textwidth]{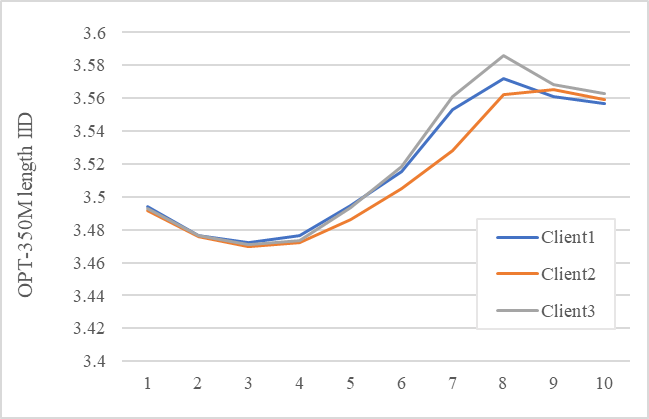}
        \caption{Test loss for OPT-350M with IID data. \(E\) = 3.}
        \label{fig:l22}
    \end{subfigure}

    \begin{subfigure}[b]{0.2\textwidth}
        \includegraphics[width=\textwidth]{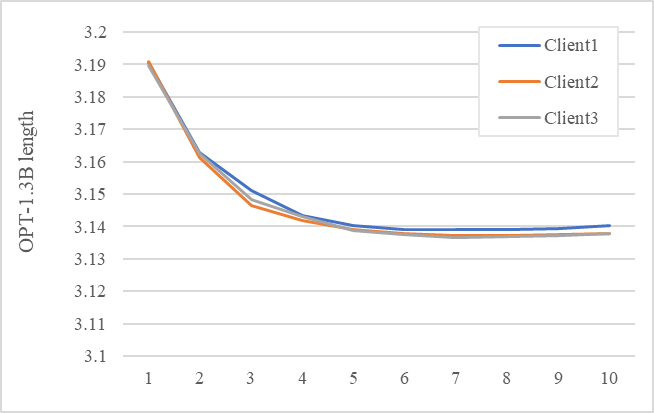}
        \caption{Test loss for OPT-1.3B with non-IID data. \(E\) = 1.}
        \label{fig:l31}
    \end{subfigure}
    \hspace{0.03\textwidth}
    \begin{subfigure}[b]{0.2\textwidth}
        \includegraphics[width=\textwidth]{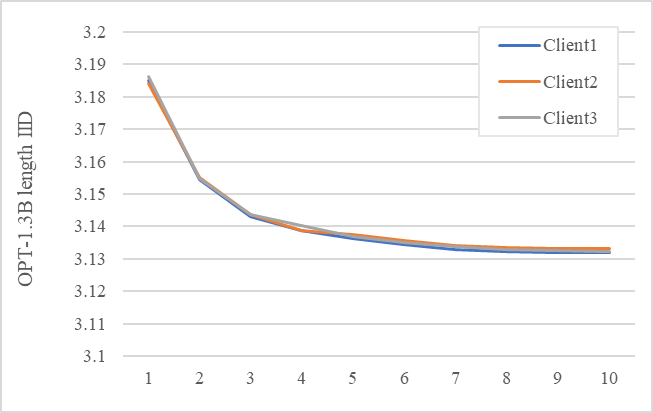}
        \caption{Test loss for OPT-1.3B with IID data. \(E\) = 1.}
        \label{fig:l32}
    \end{subfigure}

    \caption{The client test loss curves of different sizes of models in the imbalanced control text length case. X-axis: communication round; Y-axis: test loss of the aggregated model.}
    \vspace{-3pt}
    \label{fig:labels3c}
\end{figure}

We investigate the impact of text length on the local training effect on clients, using test loss as the evaluation metric and setting \(\alpha\) to 0.1 for non-IID conditions. As can be seen from Fig. \ref{fig:labels3c}, for the two models of OPT-125M and OPT-1.3B, the trends of non-IID and IID are similar; the difference is that the test loss performs slightly better than non-IID in the case of IID data distribution. The difference between the two distributions for OPT-350M is large, not only in the convergence trend but also in the final test loss. For non-IID, except for Client3, the other two rapidly overfit, while in the IID case, they are almost the same. For Fig. \ref{fig:l22}, the test curve seems to fluctuate a lot. This is because the scales of the y-axis do not correspond to each other. Actually, the test loss of the three Clients does not change a lot. The model performance fairness across different clients is also high. This suggests that the text length does not significantly impact fairness across different clients in Non-IID cases. 


\subsection{Effect of Context Quantity on Local Training}
\begin{figure}[t]
    \centering
    \begin{subfigure}[b]{0.2\textwidth}
        \includegraphics[width=\textwidth]{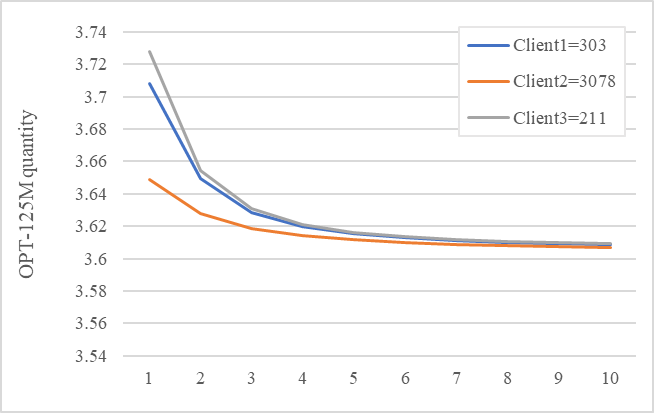}
        \caption{Test loss for OPT-125M with non-IID data. \(E\) = 3.}
        \label{fig:q11}
    \end{subfigure}
    \hspace{0.03\textwidth}
    \begin{subfigure}[b]{0.2\textwidth}
        \includegraphics[width=\textwidth]{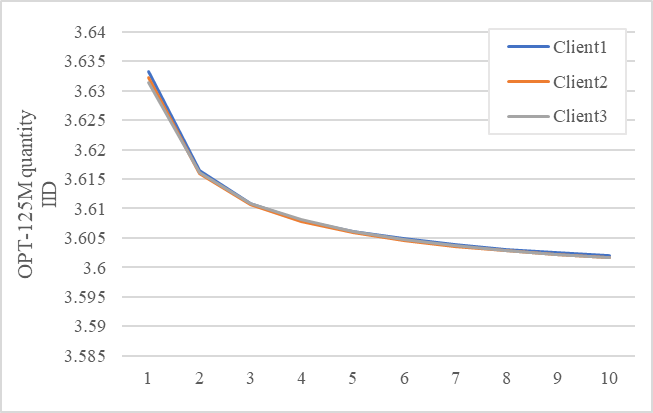}
        \caption{Test loss for OPT-125M with IID data. \(E\) = 3.}
        \label{fig:q12}
    \end{subfigure}

    \begin{subfigure}[b]{0.2\textwidth}
        \includegraphics[width=\textwidth]{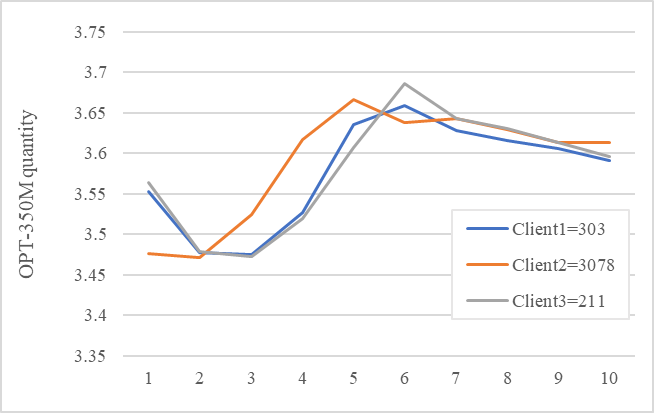}
        \caption{Test loss for OPT-350M with non-IID data. \(E\) = 3.}
        \label{fig:q21}
    \end{subfigure}
    \hspace{0.03\textwidth}
    \begin{subfigure}[b]{0.2\textwidth}
        \includegraphics[width=\textwidth]{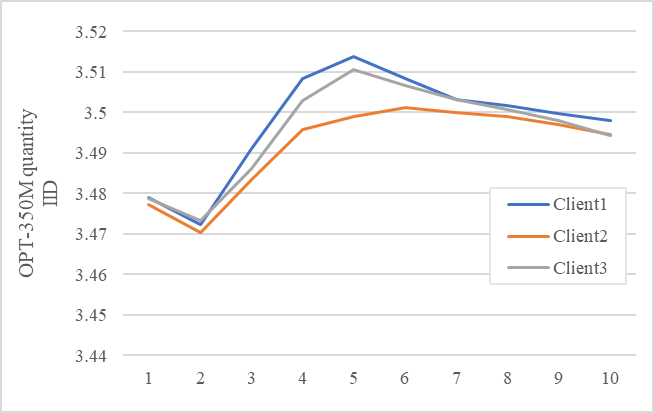}
        \caption{Test loss for OPT-350M with IID data. \(E\) = 3.}
        \label{fig:q22}
    \end{subfigure}

    \begin{subfigure}[b]{0.2\textwidth}
        \includegraphics[width=\textwidth]{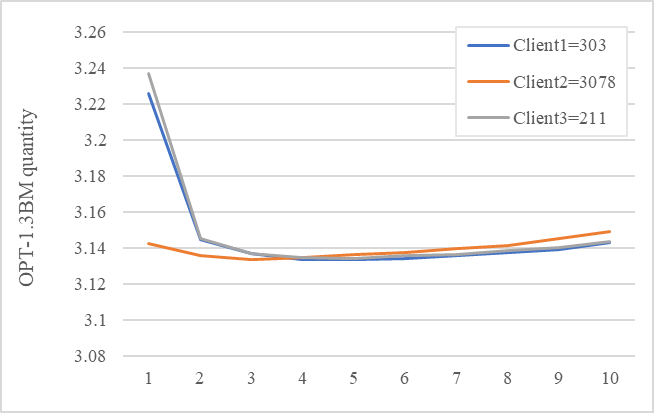}
        \caption{Test loss for OPT-1.3B with non-IID data. \(E\) = 1.}
        \label{fig:q31}
    \end{subfigure}
    \hspace{0.03\textwidth}
    \begin{subfigure}[b]{0.2\textwidth}
        \includegraphics[width=\textwidth]{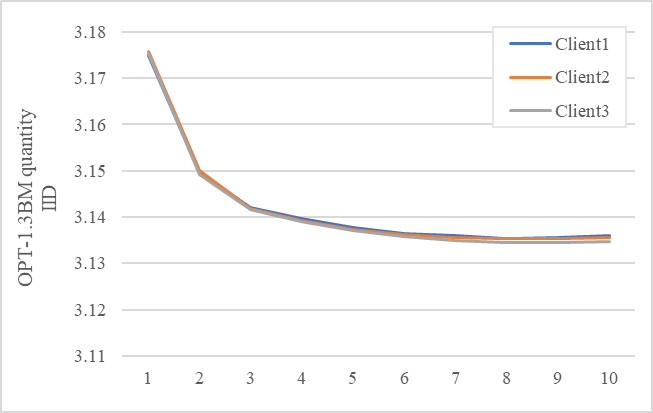}
        \caption{Test loss for OPT-1.3B with IID data. \(E\) = 1.}
        \label{fig:q32}
    \end{subfigure}

    \caption{The client test loss of different sizes of models in the imbalanced control text quantity case. X-axis: communication round; Y-axis: test loss of the aggregated model.}
    \vspace{-3pt}
    \label{fig:quantity3c}
\end{figure}

We explore the effect of text quantity on the local training effect on clients. The evaluation metric is also test loss, and the imbalance of text quantity is simulated with \(\alpha\)=0.1. From Fig. \ref{fig:quantity3c}, it can be seen that in the case of data imbalance, the starting test loss of different clients has a large gap while gradually approaching over time. Client2, which has a large amount of text data, performed well at the beginning. However, the aggregated weights may affect the two models OPT-125M and OPT-1.3B, and overfitting occurs earlier. For IID, the performance was close except for OPT-350M. After 10  communication rounds, the IID group's overall test loss is notably lower than the non-IID group's, highlighting the pronounced impact of data quantity disparity on local training outcomes. This indicates that the quantity of text highly impacts fairness across different clients in non-IID cases.

\subsection{Experiment Result Analysis}
In terms of the global model, both the length and the quantity of text have an impact on the fine-tuning results. The text length in OPT-125 and OPT-350 has a greater impact, while the experiment results of OPT-1.3B are more similar. This indicates that the small model is more sensitive to the length of the text, while the large model is more robust. The effectiveness of training is influenced by the quantity distribution of data. Optimal outcomes are achieved when the data is evenly distributed. A highly unbalanced data distribution yields marginally better results compared to a situation where the data is more evenly distributed, resulting in a parabolic shape in the quantity of training effects.

In terms of localized training, it's evident that variations in data quantity significantly influence the effect across clients. Unbalanced quantities will lower the upper bound of the model's performance. Additionally, the length of the text plays a role, not uniformly across all clients, but in terms of the model's local efficiency and fairness.

Experiment results suggest that text length and text quantity are critical factors in optimizing text generation tasks within federated learning environments, making them key indicators for reference.
\section{Conclusion}
The work employed models of varying sizes and utilized the Dirichlet distribution to obtain data with different levels of imbalance. This was compared with IID data to explore the feasibility of using the quantity and length of texts as benchmarks for non-IID text in federated learning. The results indicate that an imbalance in the number of texts not only affects local training on the client side but also impacts the effectiveness of the global model. However, the length of texts has a lesser effect on local training and a more significant impact on the global model. Thus, both the number and length of texts can serve as reference standards for non-IID text. Additionally, using LoRA significantly reduces the number of parameters, thereby decreasing communication costs and computational resource requirements. 

\bibliography{aaai25}

\end{document}